\documentclass[letterpaper,10pt,conference]{ieeeconf}

\IEEEoverridecommandlockouts
\usepackage{cite}
\usepackage{amsmath,amssymb}
\usepackage{graphicx}
\usepackage{booktabs}
\usepackage{multirow}
\usepackage{float}
\usepackage[table]{xcolor}
\usepackage{balance}
\usepackage{microtype}
\usepackage{placeins}
\newcommand{\bestresult}[1]{\ensuremath{\mathbf{#1}}}
\usepackage[hidelinks]{hyperref}
\hypersetup{pdftitle={AirwayView: Learning to Improve and Hold Decision-Sufficient Views at Airway Junctions},
  pdfauthor={Chi Kit Ng, Lintao Song, Jinsong Lin, Xinxin Liu, Ho Yin Chim, Zhiqing Tang, Tianao Ren, Sze Hang Ng, Hongliang Ren}}

\title{\LARGE \bf
AirwayView: Learning to Improve and Hold\\
Decision-Sufficient Views at Airway Junctions
}

\author{%
Chi Kit Ng\textsuperscript{1,*}, Lintao Song\textsuperscript{1,*},
Jinsong Lin\textsuperscript{1,*}, Xinxin Liu\textsuperscript{2}\\
Ho Yin Chim\textsuperscript{3}, Zhiqing Tang\textsuperscript{1},
Tianao Ren\textsuperscript{4}, Sze Hang Ng\textsuperscript{2},
Hongliang Ren\textsuperscript{1,\ensuremath{\dagger}}\\[4pt]
{\footnotesize \textsuperscript{1}Department of Electronic Engineering, The Chinese University of Hong Kong}\\
{\footnotesize \textsuperscript{2}Department of Surgery, The Chinese University of Hong Kong}\\
{\footnotesize \textsuperscript{3}National University of Singapore}\\
{\footnotesize \textsuperscript{4}Department of Mechanical Engineering, Stanford University}\\[3pt]
{\footnotesize \textsuperscript{*}Equal contribution.\quad
\textsuperscript{\ensuremath{\dagger}}Corresponding author
\href{mailto:hlren@ee.cuhk.edu.hk}{\texttt{hlren@ee.cuhk.edu.hk}}.}%
}

\begin{document}

\maketitle
\thispagestyle{plain}
\pagestyle{plain}
\clubpenalty=10000
\widowpenalty=10000
\displaywidowpenalty=10000
\brokenpenalty=10000
\finalhyphendemerits=50000

\begin{abstract}
Reliable visualization of competing airway branches can help clinicians interpret the route toward peripheral pulmonary lesions during robotic bronchoscopy. Existing navigation policies emphasize reaching or entering a branch, leaving the local adjustment and stopping needed for an informative junction view underexplored. We formulate \emph{decision-sufficient junction viewing} as a local active-perception primitive that adjusts the view between junction arrival and branch entry. AirwayView uses a causal visual policy that predicts three-axis corrective motion and an explicit sufficiency-based hold decision, with anchor-relative preference as auxiliary supervision. We further refine the policy through expert-driven collections that first revisit demonstrated junctions and then broaden the viewpoint distribution. On development replay, junction-diverse refinement improves sufficiency area under the precision--recall curve (AUPRC) from 0.44 to 0.78 and reaches a stopping $F_1$ of 0.79. In 60 policy-controlled phantom trials spanning four junctions unseen during training and three nominal starts, task success increases from 0/12 to 6/12 and 11/12 across the three training stages, while temporal behavior cloning (BC) and a Robotics Transformer 1 (RT-1) baseline achieve 0/12. Closed-loop trajectories show that successful adjustment requires corrective bending followed by appropriate stopping. These results support decision-sufficient viewpoint adjustment as a useful intermediate primitive for robotic bronchoscopy.
\end{abstract}

\section{Introduction}

Navigation and robotic bronchoscopy extend access toward peripheral pulmonary lesions, but the planned route still has to be interpreted at successive airway branches \cite{folch2019navigate,kemp2020navigation,chen2021benefit,agrawal2020robotic}. A localization system can report junction arrival while the live camera exposes only one ostium or crops a competing lumen. A physician or external planner must then choose from incomplete evidence. Centering one lumen can compound the ambiguity by keeping its sibling out of view.

This creates a perception--action gap between \emph{junction arrival} and \emph{branch entry}. Before accepting a branch-selection instruction, the robot needs a short local maneuver that exposes candidate lumens in a stable view with clear boundaries and minimal cropping.

Bronchoscopic perception methods address anatomical labeling, lumen tracking, depth, and camera localization \cite{keuth2024airway,tian2025bronchotrack,tian2024pans,tian2024ddvnb,sganga2019offsetnet}. Autonomous and shared-control systems address route execution, target-directed navigation, and local motion control \cite{zhang2024ai,zhao2024bronchocopilot,ng2026endolift,ng2025contact}. These advances answer ``where am I?'' or ``where should I travel?'' A complementary question remains at the perception--action boundary: \emph{which local camera motion will reveal a more informative junction view, and when is that view already good enough to stop moving?}

\begin{figure*}[t]
    \centering
    \includegraphics[width=0.9\textwidth]{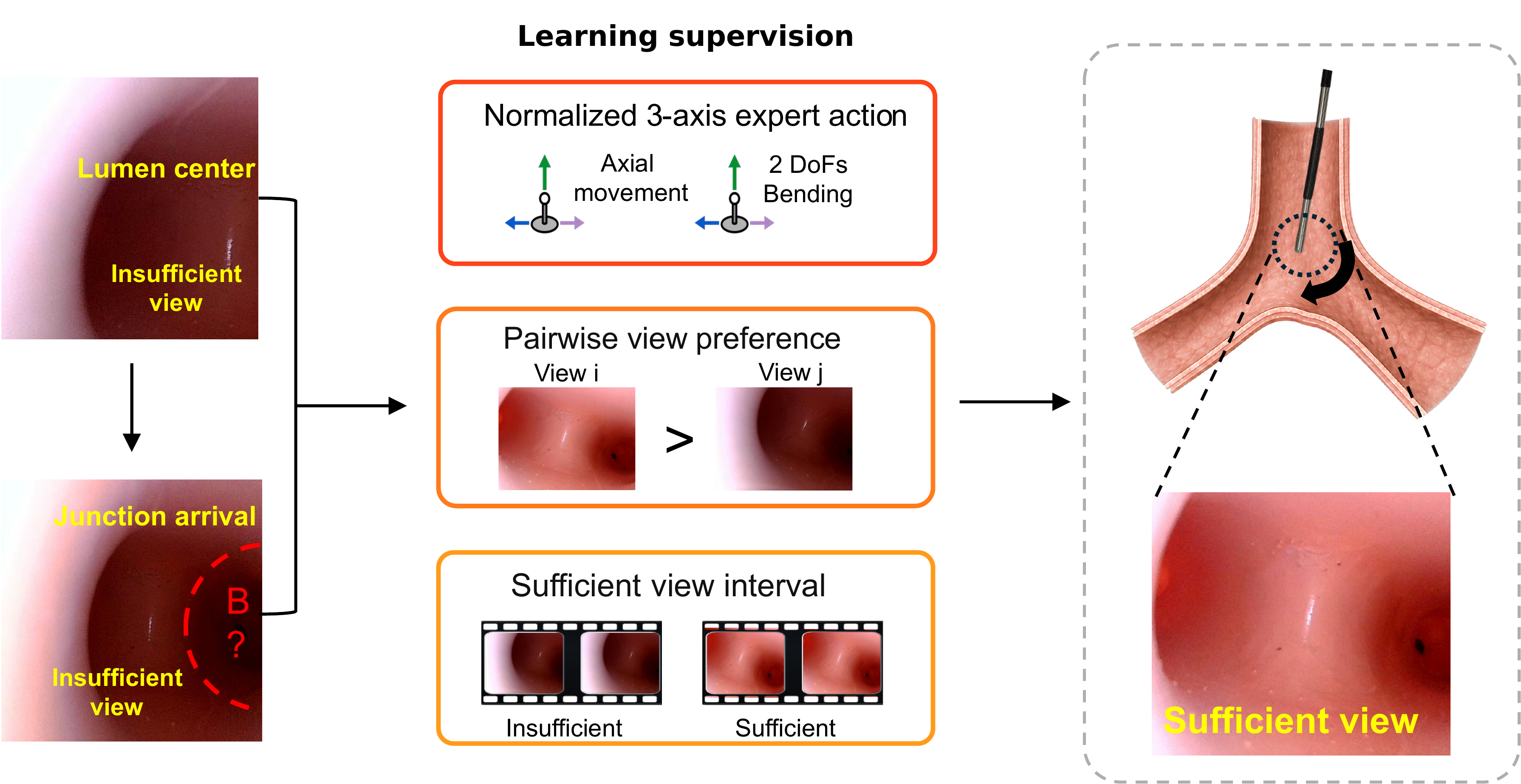}
    \caption{\textbf{Airway-junction viewpoint adjustment and its supervision.}
    A view may center one lumen or indicate junction arrival while remaining insufficient because a competing branch is cropped or visually ambiguous. AirwayView learns the local correction from three complementary annotations: expert action in three degrees of freedom (DoFs), pairwise view preference, and sufficient-view intervals. The resulting primitive adjusts the camera composition toward a view that exposes the candidate branches for selection.}
    \label{fig:problem}
\end{figure*}

We formulate this missing step as \emph{airway-junction viewpoint adjustment}, an active-perception primitive positioned between junction arrival and target-branch entry (Fig.~\ref{fig:problem}) \cite{bajcsy1988active}. Reaching, centering, and decision sufficiency are therefore separate objectives: reaching establishes proximity, centering optimizes one opening, and sufficiency makes branch choice possible.

The distinction between \emph{better} and \emph{sufficient} is central. A preferred image can still omit a candidate lumen, while an absolute sufficiency label gives little guidance about how to improve an inadequate view. These labels define complementary targets: improvement direction and acceptance of the resulting composition.

AirwayView learns from expert corrections, view preferences relative to the episode anchor, and intervals labeled sufficient. A shared temporal state predicts a three-axis correction and re-evaluates the hold decision. Demonstration-matched refinement (DMR) revisits the training junctions. Junction-diverse refinement (JDR) adds layouts with randomized initial views.

The contributions are:
\begin{itemize}
\item \textbf{Decision-sufficient junction viewing:} We formulate viewpoint adjustment as a local active-perception primitive between junction arrival and branch entry.
\item \textbf{Decoupled correction and stopping:} AirwayView learns corrective motion together with an explicit sufficiency decision, using anchor-relative preference as auxiliary supervision for view improvement.
\item \textbf{Distribution-targeted physical refinement:} Expert-driven collection broadens viewpoint coverage and raises closed-loop success from 0/12 to 6/12 and 11/12 on four held-out junctions.
\end{itemize}

\section{Related Work}

\subsection{Navigation, localization, and airway understanding}

Navigation bronchoscopy combines planning, registration, and intraoperative guidance \cite{kemp2020navigation,folch2019navigate,giri2022advancements}. Virtual navigation and manual bronchial branch tracing offer complementary ways to plan access to peripheral lesions \cite{giri2021virtual,kho2024bronchial}. Robotic platforms improve reach and articulation, with clinical performance depending on imaging confirmation and workflow \cite{chen2021benefit,agrawal2020robotic,zhang2024robotic,kops2023diagnostic,abdelghani2024conebeam}. Temporal anatomical labeling identifies branches \cite{keuth2024airway}, and BronchoTrack combines lumen detection, tracking, and airway association \cite{tian2025bronchotrack}. Visual registration and learned pose estimation localize the bronchoscope \cite{tian2024pans,tian2024ddvnb,sganga2019offsetnet}. Three-dimensional reconstruction from monocular endoscopic video can provide geometric context for navigation \cite{yang20243d,guo2025endo3r}. A systematic review documents broader uses of artificial intelligence in airway navigation and image interpretation \cite{cold2025artificial}. Active view composition exposes the alternatives for branch selection.

\subsection{Autonomous control and active perception}

Prior work on endoscopic control spans shared-control steering \cite{zhang2024ai}, multimodal reinforcement learning in simulation \cite{zhao2024bronchocopilot}, and pure-vision or language-conditioned policies \cite{wu2026long,kit2025endovla,11777392}. EndoVLA-OOV uses language and temporal visual context to bring previously observed gastrointestinal targets back into view \cite{ng2026endovla}. EndoLIFT uses language to disambiguate advancement and retraction when the views resemble one another \cite{ng2026endolift}. Tian et al.\ use local exploration during training to estimate a deformation Jacobian \cite{tian2025jacobian}, and EndoWAM grounds action prediction in estimates of task-relevant regions in future observations \cite{lin2026endowam}. More general visuomotor policies model actions through direct regression, action chunking with Transformers (ACT), diffusion, or discrete tokens \cite{Zhao-RSS-23,chi2025diffusion,brohan2022rt}. We implement these action-modelling mechanisms under a common encoder and data protocol for junction-view adjustment. At a junction, AirwayView learns to correct the camera view and assess whether it shows enough to select a branch.

\subsection{Preference-guided and iterative robot learning}

Pairwise feedback is widely used when a desired behavior is easier to compare than to encode as a scalar reward \cite{christiano2017deep}. Ranking demonstrations can reveal latent task quality and support reward learning beyond direct imitation \cite{brown2019extrapolating}. However, preference defines an ordering rather than an acceptance boundary: a view may be better than its anchor yet still omit a branch. AirwayView therefore learns preference and sufficiency as separate outputs. EndoRSI uses failed physical rollouts to guide policy revision under human oversight \cite{ng2026endorsi}. AirwayView's staged refinement is related to dataset aggregation, which addresses state-distribution shift in sequential policies \cite{ross2011reduction}. In the reported rounds, the expert drives while policy proposals are recorded in shadow, and updates occur only between collections.

\begin{figure}[t]
    \centering
    \includegraphics[width=0.88\columnwidth]{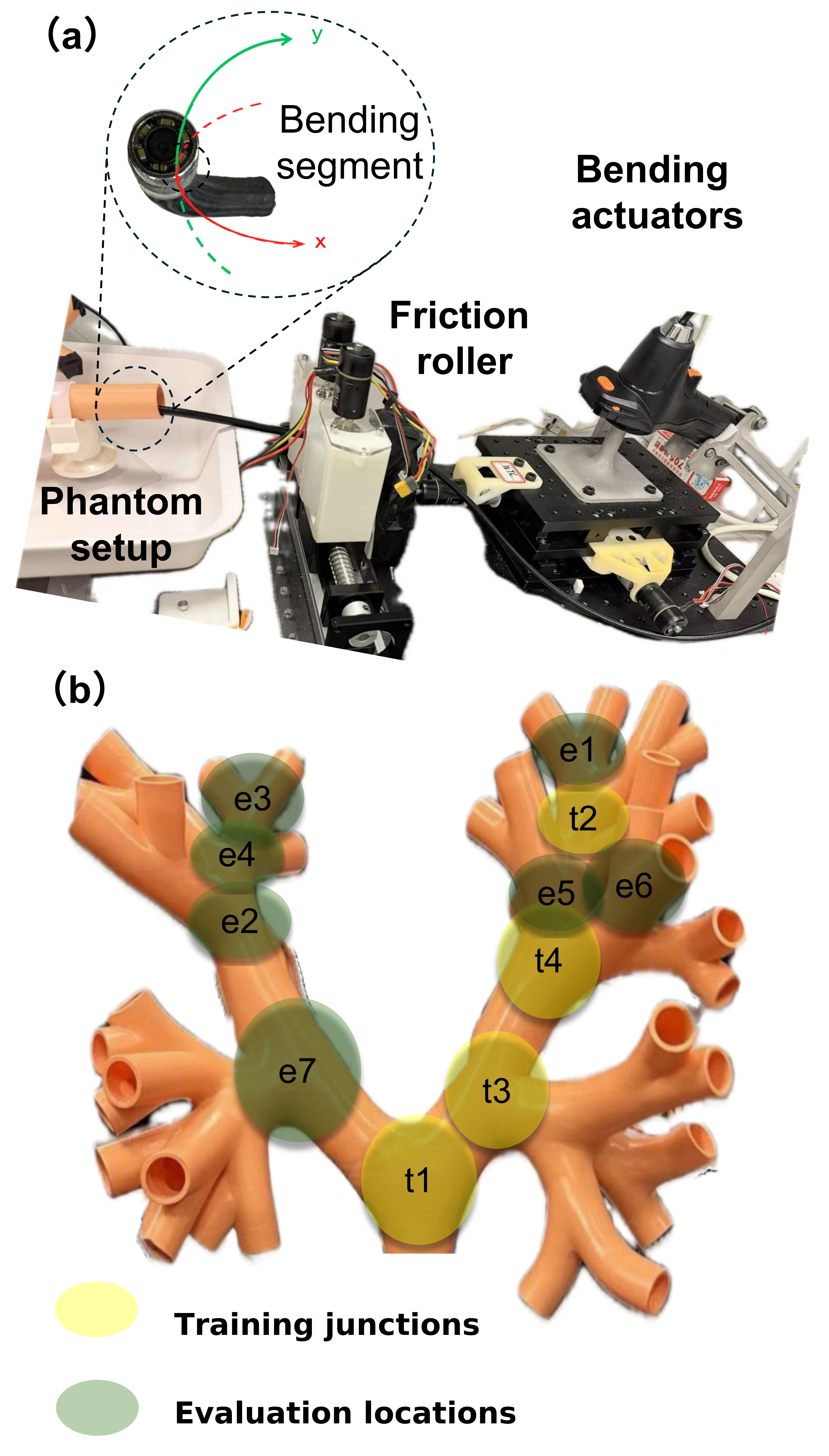}
    \caption{\textbf{Robotic platform and physical evaluation.}
    \textbf{a}, Motorized feeding and two-axis bending provide the three learned control axes; the camera view is adjusted through the feeding and distal-bending mechanisms.
    \textbf{b}, The supplied airway-phantom map identifies training junctions t1--t4 and evaluation locations e1--e7. The robot cohort uses four junctions unseen during training, e4--e7, with S01--S03 nominal starts and six policy/reference groups (72 trials). }
    \label{fig:setup}
\end{figure}

\section{Robotic Platform and Evaluation Setup}
\label{sec:setup}

The platform combines a motorized feeding mechanism for insertion/retraction with two-axis distal bending and an endoscopic red--green--blue (RGB) camera (Fig.~\ref{fig:setup}a). The learned action controls insertion, up--down (UD) bending, and left--right (LR) bending through a serial motor interface. The camera supplies $480\times480$ circularly masked views; the policy resizes images to $224\times224$. Physical evaluation uses a nominal 10-Hz observe--infer--dispatch loop. 

The phantom task is local viewpoint adjustment. The operator positions the scope at a nominal start, initiates execution, and can manually end a trial or request a safety stop. Model trials otherwise end after ten consecutive policy-hold ticks or a 30-s limit. Expert reference uses manual control; the cohort and blinded success assessment are specified in Sec.~\ref{sec:experiments}.

\begin{figure*}[!t]
    \centering
    \includegraphics[width=0.99\textwidth]{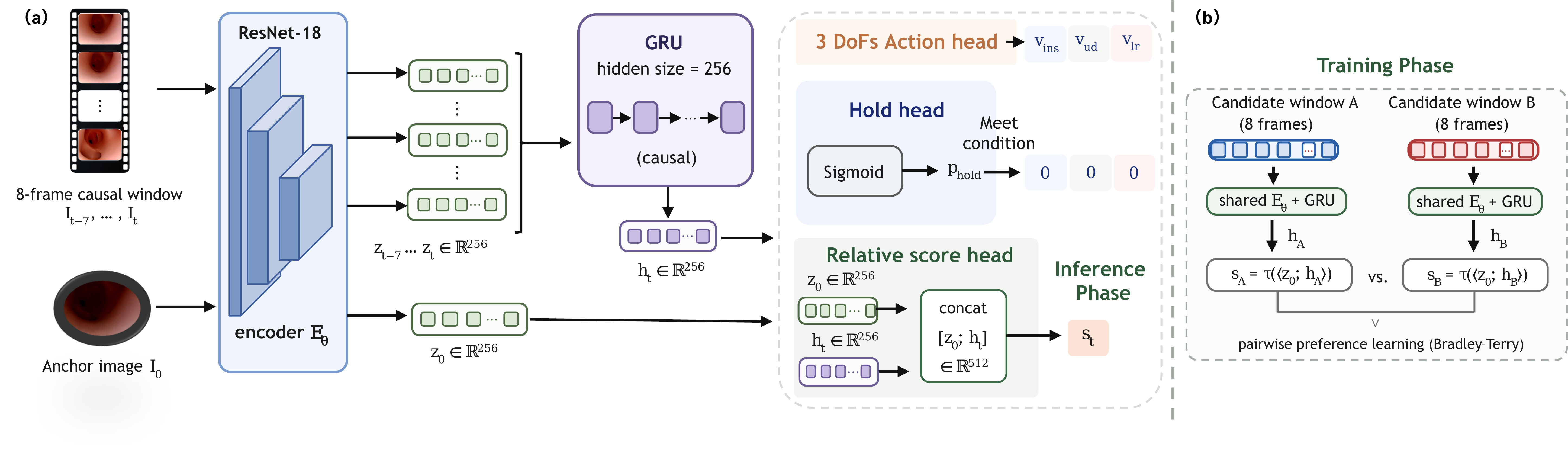}
    \caption{\textbf{AirwayView architecture and preference supervision.}
    \textbf{a}, Eight causal images share a ResNet-18 encoder and a $512\!\rightarrow\!256$ projection, followed by a 256-dimensional GRU. The temporal state predicts correction and hold; only the auxiliary score head also uses the reset-anchor feature. The runtime applies the hold condition before safety-filtered dispatch.
    \textbf{b}, Two candidate windows share the encoder, GRU, and anchor-conditioned score head. Their score difference defines the Bradley--Terry ranking loss; action and hold labels use separate optimizer steps.}
    \label{fig:architecture}
\end{figure*}

\section{Method}

\subsection{Observation, action, and robot interface}

The policy input is a window of $W$ ordered RGB images:
\begin{equation}
    \mathcal{O}_t=\{I_{t-W+1},\ldots,I_t\}.
\end{equation}
Missing initial context repeats the earliest frame. The reset image $I_0$ is retained separately as the anchor, outside the gated recurrent unit (GRU) sequence. The canonical action order is
\begin{equation}
    \mathbf{a}_t=[v_{\mathrm{ins}},v_{\mathrm{ud}},v_{\mathrm{lr}}]^\top\in[-1,1]^3,
    \label{eq:action}
\end{equation}
where magnitude is normalized speed and sign specifies direction. Clipping, deadbanding, and safety rate limiting precede motor dispatch:
\begin{equation}
\mathbf m_t=
[-\alpha_{\rm ins}v_{\rm ins},-\alpha_{\rm ins}v_{\rm ins},
\alpha_{\rm ud}v_{\rm ud},\alpha_{\rm lr}v_{\rm lr}]^\top ,
\label{eq:dispatch}
\end{equation}
where $\mathbf m_t$ contains coupled insertion motors M1/M2 and bending motors M3/M4. A hold requests zero on all three learned axes. Stale observations or hardware faults trigger a fail-closed zero command and termination. Logs store requested and safety-filtered commands.

\subsection{Three-head temporal policy}

We normalize each image to $[0,1]$ and resize it to $224\times224$.
A shared 18-layer residual network (ResNet-18) without its classifier encodes the image.
The rectified linear unit~(ReLU) follows a shared $512\!\rightarrow\!256$ projection:
\begin{equation}
    \mathbf z_k=\rho\!\left(W_p\,\operatorname{ResNet18}(I_k)\right)\in\mathbb R^{256}.
\end{equation}
Stage~I initializes the ResNet-18 with \texttt{weights=None}; DMR and JDR load
the selected parent checkpoint rather than reinitializing the encoder.
A one-layer GRU maps the ordered window features to
\begin{equation}
    \mathbf h_t=\operatorname{GRU}(\mathbf z_{t-W+1},\ldots,\mathbf z_t)\in\mathbb R^{256}.
\end{equation}
The shared encoder maps $I_0$ to $\mathbf z_0$. In a single-window pass, $\tilde{\mathbf h}_t=\operatorname{Drop}_{0.2}(\mathbf h_t)$ is shared by all enabled heads:
\begin{equation}
\begin{aligned}
\hat{\mathbf a}_t&=\tanh a(\tilde{\mathbf h}_t),\\
\hat p_t&=\sigma(q(\tilde{\mathbf h}_t)),\\
\hat s_t&=\tau([\mathbf z_0;\tilde{\mathbf h}_t]).
\end{aligned}
\end{equation}
The action, hold, and score multilayer perceptrons (MLPs) have widths $256\!\rightarrow\!128\!\rightarrow\!3$, $256\!\rightarrow\!64\!\rightarrow\!1$, and $512\!\rightarrow\!128\!\rightarrow\!1$, respectively. The action head adds ReLU, 0.2 dropout, and $\tanh$; evaluation disables dropout. For preference training, two candidate windows share the encoder, GRU, and score head, yielding anchor-conditioned $\hat s_A$ and $\hat s_B$. At deployment, correction and hold depend on the temporal state; the anchor enters only the auxiliary score head (Fig.~\ref{fig:architecture}).

At inference, the hold logit is converted into motion suppression by a configurable threshold-and-persistence rule. Let $\tau_h$ be the threshold and $K\geq1$ the required number of consecutive positive estimates. The policy-requested command before safety filtering is
\begin{equation}
\mathbf{a}^{\mathrm{exec}}_t=
\begin{cases}
\mathbf{0}, & \hat{p}_{t-j}\ge \tau_h\quad\forall j\in\{0,\ldots,K-1\},\\
\hat{\mathbf a}_t, & \text{otherwise}.
\end{cases}
\label{eq:hold}
\end{equation}
Replay uses $\tau_h=0.1$. Physical evaluation uses 0.15 for Stage~I and 0.1 for DMR/JDR, with $K=1$ throughout. We select hold thresholds on development replay and freeze them before physical evaluation. The runtime recomputes Eq.~\eqref{eq:hold} at each frame. Physical dwell counts policy-hold decisions. Logs track any rate-limited deceleration tick as a separate event.

\subsection{Preference, sufficiency, and losses}
\label{sec:losses}

Each preference example contains an episode anchor, two causal windows, and a preferred-final-view label. Presentation order is randomized; validation balances the displayed winner. Ties and uncertain judgments retain unknown or optional tie targets.

Sufficiency is annotated independently as intervals. Let $z_t\in\{0,1\}$ denote an insufficient/sufficient target and $m_t\in\{0,1\}$ indicate label availability; uncertain boundaries can have $m_t=0$. A preferred candidate may remain insufficient (Fig.~\ref{fig:problem}). Recorded expert actions are preserved even when $z_t=1$, so the action and hold heads learn distinct targets. The runtime supervisor in Eq.~\eqref{eq:hold} converts sufficiency into a zero-action request.

The action loss $\mathcal{L}_{\mathrm{act}}$ is Huber loss on expert velocity, and the sufficiency loss $\mathcal{L}_{\mathrm{hold}}$ is masked binary cross entropy. For a decisive pair with $y_{ij}=1$ when window $j$ is preferred to window $i$,
\begin{equation}
\mathcal{L}_{\mathrm{rank}}=-y_{ij}\log\sigma(\hat{s}_j-\hat{s}_i)
-(1-y_{ij})\log\sigma(\hat{s}_i-\hat{s}_j).
\end{equation}
The action, pair, and hold samplers are independent. If $b\in\{\mathrm{act},\mathrm{rank},\mathrm{hold}\}$ denotes the loader selected for the current step, the implementation updates
\begin{equation}
\theta\leftarrow\theta-\eta\nabla_\theta\!\left(\lambda_b\mathcal{L}_b\right),
\qquad [\lambda_{\mathrm{act}},\lambda_{\mathrm{rank}},\lambda_{\mathrm{hold}}]=[1,0.2,0.2].
\end{equation}
Each training batch supplies the target type of its selected loader. Checkpoint selection minimizes the development-only weighted sum of the three available validation losses; ranking and hold metrics are reported for interpretation, and the hold threshold and dwell are calibrated only on development data.

\subsection{Staged viewpoint-distribution refinement}

The three stages retain the same policy interface and differ in the viewpoint distribution used for supervision. DMR rounds 0 and 1 contain 53 episodes/5,235 frames and 52 episodes/4,894 frames, respectively; JDR round 0 adds 52 episodes/5,847 frames.

\textbf{Stage I---demonstration initialization:} train from scratch on recorded expert episodes to learn the basic correspondence among appearance, correction, preference, and sufficiency. The resulting checkpoint is denoted $\theta^{I}$.

\textbf{Stage II---DMR:} gather additional expert trajectories at t1/t2, the same physical training junctions used for the demonstration distribution, while policy proposals are logged in shadow. Human reviewers label sufficient-view intervals. At round $k$, the training set combines the demonstrations with the reviewed collections:
\begin{equation}
    D^{II}_k=D^I\cup\bigcup_{r=0}^{k} C^{II}_r.
\end{equation}
Weights are updated between collection rounds on a controlled mixture of demonstrations and the new DMR data. The selected round-0 checkpoint samples action and hold examples with probability 0.5 from $D^I$ and 0.5 from $C^{II}_0$. Ranking steps replay the original $D^I$ preference pairs, while the new collection supplies action and sufficiency labels. This stage tests whether revisiting the demonstrated junction family repairs errors caused by imperfect arrival states.

\textbf{Stage III---JDR:} initialize from the selected Stage~II checkpoint and repeat the collect--aggregate--update cycle at t3/t4 under randomized initial views. The refinement junctions belong to the training set; e4--e7 are four junctions unseen during training. The selected mixture samples 0.50 from $D^I$, 0.125 from each of the two DMR rounds, and 0.25 from JDR round~0. The JDR collection supplies actions and sufficiency labels. Preference training keeps the original $D^I$ pairs.

\section{Experiments}
\label{sec:experiments}

\subsection{Protocol and evaluation questions}

Our experiments follow the transition from imitating expert images to acquiring a useful view on the robot. We first ask whether explicit sufficiency identifies when a view is ready for branch selection. We then test whether expanding the training-view distribution changes closed-loop behavior, and examine the trajectories behind the resulting task successes and failures.

The demonstration corpus contains 26 episodes, 7,649 frames, and 769 preference pairs across 11 junctions. Training uses t1/t2 for demonstration-matched collection and t3/t4 for junction-diverse collection. Development replay uses e1 (604 frames) and e2 (616 frames). Of 3,547 reviewed frames in the corpus, 1,310 are sufficient; another 4,102 frames have unknown labels. DMR and JDR checkpoint selection uses e1 and e2, respectively. The reported replay scores characterize these development distributions; physical evaluation uses four junctions unseen during training (e4--e7).

All learned methods share the ResNet-18 encoder and image transforms. AirwayView uses a 256-dimensional GRU, eight causal $224\times224$ frames approximately 100 ms apart, and action, ranking, and hold loss weights of 1, 0.2, and 0.2. Stage~I uses adaptive moment estimation with decoupled weight decay (AdamW) for up to 20 epochs (batch 16, learning rate $10^{-4}$, weight decay $10^{-5}$, patience five); refinement uses $2\times10^{-5}$. Baseline replay results average three seeds; AirwayView uses the selected stage checkpoints. Replay uses hold threshold $\tau_h=0.1$ and persistence $K=1$.

Physical evaluation comprises 60 policy-controlled trials and 12 expert-reference trials: three AirwayView stages, temporal BC, RT-1, and an expert are evaluated on four junctions unseen during training with three nominal starts, S01--S03. An independent physician, blinded to the policy identity, assigns task success from the recorded videos. Success requires a junction composition suitable for branch selection. The expert achieves 12/12 and provides an expert reference. Task outcome is assessed separately from controller dwell, timeout, manual end, and manual safety stop. The operator resets the scope to the nominal arrival pose before every trial.

\subsection{Explicit sufficiency identifies a usable stopping point}

The first experiment tests the distinction between a small action and a sufficient view. We compare explicit hold prediction with a deadband-based stopping rule applied to BC, temporal BC, ACT, a Diffusion model, and RT-1 (Table~\ref{tab:offline-stopping}). The learned baselines implement these action mechanisms with the same encoder and dataset. Their stop rule activates when all requested axes lie within the dispatch deadband. Yet expert actions lie in that deadband on only 18/74 sufficient e1 frames and 35/74 sufficient e2 frames. This gap motivates learning when to stop from labeled views.

On e1, DMR reaches stopping $F_1=0.511$, close to temporal BC at 0.508. On junction-diverse e2, JDR reaches 0.787, exceeding the strongest action-derived result of 0.542 and the aggregation-matched RT-1 result of 0.524. Explicit sufficiency is therefore most useful in the more diverse replay setting, where the magnitude of a predicted correction is a weaker indicator of visual readiness. The two auxiliary heads add only approximately $8.2\times10^4$ parameters (0.70\%) to the action backbone; AirwayView uses 12 million effective parameters.

\begin{table}[t]
\caption{\textbf{Offline stopping $F_1$.} D: demonstrations; D+S: demonstrations plus shadow aggregation. AirwayView uses the e1-selected DMR and e2-selected JDR checkpoints. Baselines are three-seed means. Bold marks the best result in each split.}
\label{tab:offline-stopping}
\centering\scriptsize
\setlength{\tabcolsep}{4pt}
\begin{tabular}{llrr}
\toprule
Model & Data & e1 $\uparrow$ & e2 $\uparrow$\\
\midrule
Always stop & -- & 0.354 & 0.481\\
BC ($W=1$) \cite{pomerleau1988alvinn} & D & 0.433 & 0.303\\
Temporal BC ($W=8$) \cite{pomerleau1988alvinn} & D & 0.508 & 0.386\\
ACT \cite{Zhao-RSS-23} & D & 0.354 & 0.481\\
ACT \cite{Zhao-RSS-23} & D+S & 0.354 & 0.456\\
Diffusion model \cite{chi2025diffusion} & D & 0.221 & 0.212\\
Diffusion model \cite{chi2025diffusion} & D+S & 0.291 & 0.318\\
RT-1 \cite{brohan2022rt} & D & 0.354 & 0.542\\
RT-1 \cite{brohan2022rt} & D+S & 0.354 & 0.524\\
\midrule
AirwayView (ours) & D+S & \bestresult{0.511} & \bestresult{0.787}\\
\bottomrule
\end{tabular}
\end{table}

\subsection{Broader training views enable physical adjustment}
\label{sec:real-results}

Stage~I learns from demonstrations, but all 12 robot trials fail. Its characteristic behavior is forward advance with ineffective bending: the scope enters an arbitrary lumen and leaves the junction before acquiring a useful composition. This failure exposes the central transfer gap. Actions learned from recorded expert views may fail on arrival views generated by the policy itself.

DMR first revisits the demonstrated junction family under additional arrival views. Success rises to 6/12, with all six successes at e6/e7. The corresponding e1 ranking loss decreases from 0.90 to 0.59 (Table~\ref{tab:matched-stages}). These results show that additional expert correction can turn the initial policy into a useful local behavior, while the failures at e4/e5 reveal the remaining coverage gap.

JDR broadens collection to t3/t4 with randomized starting views. On e2 development replay, sufficiency AUPRC rises from 0.44 to 0.78 and ranking loss decreases from 0.30 to 0.054. On the robot, success increases to 11/12 (92\%): JDR retains the six e6/e7 successes and adds five successes at e4/e5. The gain was not confined to one nominal arrival condition: JDR achieved 4/4, 4/4, and 3/4 successes at S01--S03, respectively. The 42-percentage-point increase over DMR is accompanied by broader spatial coverage of successful adjustment. This progression supports distribution-targeted refinement as a practical way to acquire the intended behavior at junctions unseen during training.

\begin{table}[t]
\caption{\textbf{From replay to real-robot success.} Offline entries report e1/e2 in that order within each cell; both are development splits. Real-robot success uses 12 trials per checkpoint on e4--e7. Bold marks the best value for each split or endpoint before rounding.}
\label{tab:matched-stages}
\centering\scriptsize
\setlength{\tabcolsep}{3.2pt}
\begin{tabular}{lccc}
\toprule
& \multicolumn{2}{c}{Offline} & Real robot\\
\cmidrule(lr){2-3}\cmidrule(lr){4-4}
Stage & Rank loss $\downarrow$ & AUPRC $\uparrow$ & Success $\uparrow$\\
\midrule
I & 0.90/0.54 & 0.12/0.56 & 0/12\\
I+DMR & \bestresult{0.59}/0.30 & \bestresult{0.21}/0.44 & 6/12\\
I+DMR+JDR & 1.0/\bestresult{0.054} & 0.12/\bestresult{0.78} & \bestresult{11/12}\\
\bottomrule
\end{tabular}
\end{table}

The replay results also explain what the refinement changes. JDR specializes toward the diverse distribution: e1 performance decreases while e2 improves. The new rounds supply expert actions and sufficient-view intervals, while preference supervision continues to use the original pairs. The result is a distribution-dependent change in correction and stopping through the shared representation. Table~\ref{tab:matched-stages} therefore shows that offline demonstration-matched performance is an incomplete proxy for closed-loop utility.

\begin{figure}[t]
\centering
\includegraphics[width=\columnwidth]{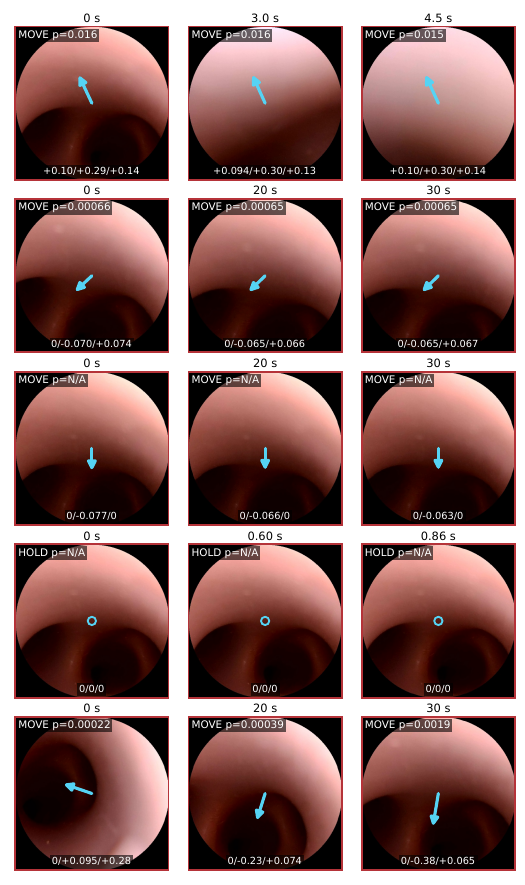}
\caption{\textbf{Representative closed-loop failures.} Rows from top: Stage~I, DMR, temporal BC, RT-1 at e4/S01, and JDR at e4/S03. Three chronological frames per trial show junction departure, insufficient correction, slow adjustment, zero-action dwell, and a difficult-start failure (supplementary video of JDR at e4 S03), respectively. Stage~I ends manually; DMR, BC, and JDR time out; RT-1 completes its dwell. Overlays show time, executed insertion/UD/LR command, hold probability (not applicable, N/A, for baselines), and MOVE/HOLD. Arrows depict signed bending commands; circles mark zero.}
\label{fig:real-failures}
\end{figure}

\begin{figure*}[t]
\centering
\includegraphics[width=\textwidth]{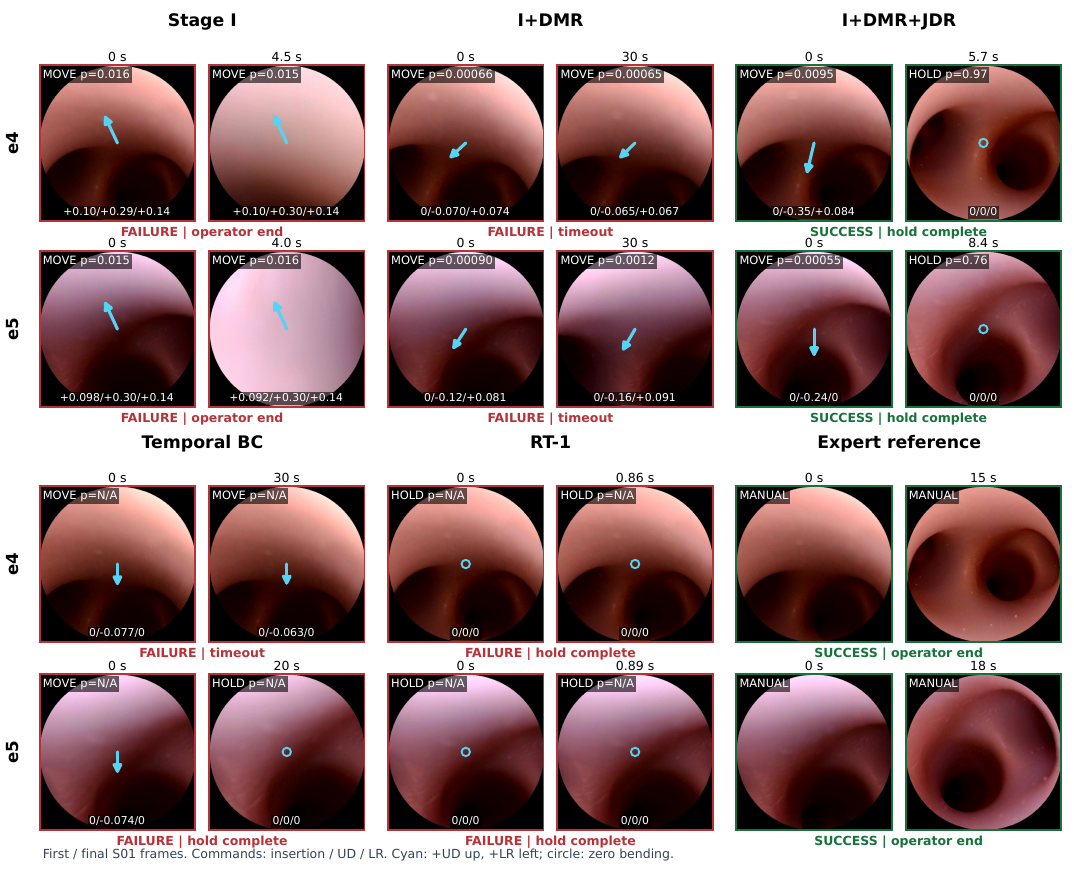}
\caption{\textbf{Real-robot demonstrations at e4 and e5.} The upper two rows compare Stage~I, DMR, and JDR; the lower two compare temporal BC, RT-1, and Expert reference. Each pair shows first/final S01 frames. Policy overlays report time, hold probability $p_h$ (N/A for baselines), MOVE/HOLD, and executed command $\mathbf u=(u_{\rm ins},u_{\rm ud},u_{\rm lr})$. The cyan arrow depicts bending in schematic command coordinates; a circle denotes zero. Expert frames show MANUAL control. JDR corrects the junction composition and holds, while the expert provides a successful reference.}
\label{fig:real-demo}
\end{figure*}

\subsection{Corrective motion and stopping}

Table~\ref{tab:real-outcomes} compares the policies with the expert reference. Temporal BC achieves 0/12: its adjustments are slow or ineffective. RT-1 also achieves 0/12 and produces zero executed commands across all 120 logged control rows. It completes every controller dwell while retaining an inadequate view. These failures identify two distinct requirements: sufficient corrective motion to change the junction composition, followed by holding when that composition becomes useful.

\begin{table}[t]
\caption{\textbf{Real-robot task performance ($n=12$ per group).} View success rate (SR) uses blinded physician assessment. $T$: median original-video duration among successful trials. Inference: median of per-trial median prediction time. Bold marks the best autonomous values and highlights expert success and failure counts separately.}
\label{tab:real-outcomes}
\centering\scriptsize
\setlength{\tabcolsep}{3.5pt}
\begin{tabular}{lrrrr}
\toprule
Method & SR (\%) $\uparrow$ & Fail $\downarrow$ & $T$ (s) $\downarrow$ & Infer. (ms) $\downarrow$\\
\midrule
AirwayView-I & 0 & 12 & -- & 16.5\\
I+DMR & 50 & 6 & 11 & 16.6\\
I+DMR+JDR & \bestresult{92} & \bestresult{1} & \bestresult{6.8} & 16.5\\
Temporal BC & 0 & 12 & -- & \bestresult{12.8}\\
RT-1 & 0 & 12 & -- & 13.5\\
\midrule
Expert reference & \bestresult{100} & \bestresult{0} & 16 & --\\
\bottomrule
\end{tabular}
\end{table}

Successful-trial video duration has median 6.8 s for JDR ($n=11$), 11 s for DMR ($n=6$), and 16 s for the expert ($n=12$); each method contributes its successful subset. JDR's median inference time of 16.5 ms supports the nominal 10-Hz loop. The expert's 100\% success establishes task feasibility, while JDR's 92\% locates the remaining gap. Together, these results support lightweight causal viewpoint adjustment.

\textbf{View success and successful hold completion.} JDR acquired a physician-rated decision-sufficient view in 11/12 trials; 10/12 additionally satisfied the autonomous hold-completion criterion, while one successful view at e5/S03 reached the 30-s timeout. These endpoints distinguish view acquisition from autonomous finish detection. Stage~I requires manual termination after junction departure in ten trials and a manual safety stop at e5/S02. The latter is a protective operator response to continued advancement after wall contact.

\subsection{Real-robot demonstrations}

Figure~\ref{fig:real-demo} shows e4/e5 examples from all six groups. JDR changes from corrective bending to HOLD while exposing adjacent openings; Stage~I remains in MOVE and ends with a wall-dominated image. Temporal BC changes the composition slowly, whereas RT-1 stays at zero action. It lacks fine-grained spatial snesitivity to micro-scale endoscopic lumen adjustments in pure-vision control tokens. Expert reference shows the intended composed view. The contrast connects the success progression to its physical meaning: useful correction must precede acceptance of the junction view.

Figure~\ref{fig:real-failures} follows five unsuccessful trials, including JDR's e4/S03 failure. Their distinct endpoints reinforce the separation between view quality and a controller dwell. The accompanying video shows representative closed-loop failures and successes, including Stage-I junction departure, RT-1 zero-action dwell, the manual safety stop, and JDR correction followed by hold.

\section{Discussion}

The principal result is a change in closed-loop behavior as the training-view distribution expands. Stage~I advances through the junction, DMR acquires usable views at a subset of locations, and JDR extends successful adjustment across four junctions unseen during training. This progression suggests that the missing ingredient after offline imitation is coverage of the correction states surrounding junction arrival. Expert-driven physical collection supplies those states while preserving a common policy architecture and control interface. The resulting primitive creates a useful pause between reaching an anatomical location and committing to a branch.

The stopping comparisons clarify why viewpoint adjustment deserves its own objective. A small command can reflect an adequate view, slow regression, or action collapse. Explicit sufficiency gives the policy a visual acceptance criterion, while action supervision teaches the correction needed to reach it and anchor-relative preference provides a view-improvement signal. The strongest replay separation occurs on junction-diverse views, and the robot examples show corrective bending followed by holding. The successful timeout further motivates calibrating completion after a useful view has been reached. A future closed-loop head ablation can quantify each supervision signal's individual contribution.

Development replay also reveals a mismatch between retention on the demonstration-matched distribution and closed-loop utility. JDR performs worse on e1 while improving substantially on the junction-diverse e2 replay and on the physical task. This suggests that demonstration-matched offline accuracy alone is an incomplete proxy for policy usefulness when deployment induces new viewpoint distributions. The study provides phantom evidence from four junctions unseen during training and three nominal starts, with blinded assessment by one independent physician. The current results is a necessary geometrical stepping stone before in-vivo validation. Broader validation should include more phantoms, animal study with repeatable pose resets, and additional raters. Video durations summarize recordings, and command vectors characterize control effort; frame-level sufficiency onsets and tip-motion sensing would support more direct timing and motion analyses. Biological-tissue evaluation and downstream branch-selection studies are the next steps toward establishing clinical benefit.

\section{Conclusion}

AirwayView formulates decision-sufficient junction viewing as a local active-perception primitive between junction arrival and branch entry. A causal policy combines corrective motion with an explicit sufficiency decision and auxiliary anchor-relative preference. Expert-driven refinement expands viewpoint coverage and raises real-robot task success from 0/12 to 6/12 and 11/12 at four junctions unseen during training. The trajectories show a shift from ineffective advance to corrective bending and useful holding, supporting this intermediate behavior for robotic bronchoscopy.

\section*{Acknowledgment}
OpenAI ChatGPT was used for language editing and grammatical refinement of the manuscript. All content, methodology, experiments, analyses, figures, and conclusions were developed by the authors.

\balance
\bibliographystyle{IEEEtran}
\bibliography{ref}

@article{chen2021benefit,
  title={Robotic bronchoscopy for peripheral pulmonary lesions: a multicenter pilot and feasibility study ({BENEFIT})},
  author={Chen, Alexander C and Pastis, Jr, Nicholas J and Mahajan, Amit K and Khandhar, Sandeep J and Simoff, Michael J and Machuzak, Michael S and Cicenia, Joseph and Gildea, Thomas R and Silvestri, Gerard A},
  journal={Chest},
  volume={159},
  number={2},
  pages={845--852},
  year={2021},
  publisher={Elsevier}
}

@article{folch2019navigate,
  title={Electromagnetic navigation bronchoscopy for peripheral pulmonary lesions: one-year results of the prospective, multicenter {NAVIGATE} study},
  author={Folch, Erik E and Pritchett, Michael A and Nead, Michael A and Bowling, Mark R and Murgu, Septimiu D and Krimsky, William S and Murillo, Boris A and LeMense, Gregory P and Minnich, Douglas J and Bansal, Sandeep and others},
  journal={Journal of Thoracic Oncology},
  volume={14},
  number={3},
  pages={445--458},
  year={2019},
  publisher={Elsevier}
}

@article{kemp2020navigation,
  title={Navigation bronchoscopy},
  author={Kemp, Samuel Victor},
  journal={Respiration},
  volume={99},
  number={4},
  pages={277--286},
  year={2020},
  publisher={S. Karger AG}
}

@article{agrawal2020robotic,
  title={Robotic bronchoscopy for pulmonary lesions: a review of existing technologies and clinical data},
  author={Agrawal, Abhinav and Hogarth, D Kyle and Murgu, Septimiu},
  journal={Journal of thoracic disease},
  volume={12},
  number={6},
  pages={3279--3286},
  year={2020}
}

@article{giri2022advancements,
  title={Advancements in navigational bronchoscopy for peripheral pulmonary lesions: A review with special focus on virtual bronchoscopic navigation},
  author={Giri, Mohan and Dai, Haiyun and Puri, Anju and Liao, Jiaxin and Guo, Shuliang},
  journal={Frontiers in Medicine},
  volume={9},
  pages={989184},
  year={2022},
  publisher={Frontiers Media SA}
}

@article{keuth2024airway,
  title={Airway label prediction in video bronchoscopy: capturing temporal dependencies utilizing anatomical knowledge},
  author={Keuth, Ron and Heinrich, Mattias and Eichenlaub, Martin and Himstedt, Marian},
  journal={International Journal of Computer Assisted Radiology and Surgery},
  volume={19},
  number={4},
  pages={713--721},
  year={2024},
  publisher={Springer}
}

@article{tian2025bronchotrack,
  title={{BronchoTrack}: Airway lumen tracking for branch-level bronchoscopic localization},
  author={Tian, Qingyao and Liao, Huai and Huang, Xinyan and Yang, Bingyu and Wu, Jinlin and Chen, Jian and Li, Lujie and Liu, Hongbin},
  journal={IEEE transactions on medical imaging},
  volume={44},
  number={3},
  pages={1321--1333},
  year={2025},
  publisher={IEEE}
}

@inproceedings{tian2024pans,
  title={{PANS}: Probabilistic airway navigation system for real-time robust bronchoscope localization},
  author={Tian, Qingyao and Chen, Zhen and Liao, Huai and Huang, Xinyan and Yang, Bingyu and Li, Lujie and Liu, Hongbin},
  booktitle={International Conference on Medical Image Computing and Computer-Assisted Intervention},
  pages={466--476},
  year={2024},
  organization={Springer}
}

@inproceedings{tian2024ddvnb,
  title={{DD-VNB}: A depth-based dual-loop framework for real-time visually navigated bronchoscopy},
  author={Tian, Qingyao and Liao, Huai and Huang, Xinyan and Chen, Jian and Zhang, Zihui and Yang, Bingyu and Ourselin, Sebastien and Liu, Hongbin},
  booktitle={2024 IEEE/RSJ International Conference on Intelligent Robots and Systems (IROS)},
  pages={12979--12986},
  year={2024},
  organization={IEEE}
}

@inproceedings{zhao2024bronchocopilot,
  title={BronchoCopilot: Towards autonomous robotic bronchoscopy via multimodal reinforcement learning},
  author={Zhao, Jianbo and Chen, Hao and Tian, Qingyao and Chen, Jian and Yang, Bingyu and Zhang, Zihui and Liu, Hongbin},
  booktitle={2024 IEEE/RSJ International Conference on Intelligent Robots and Systems (IROS)},
  pages={6923--6930},
  year={2024},
  organization={IEEE}
}

@article{wu2026long,
  title={Long-Short Term Agents for Pure-Vision Bronchoscopy Robotic Autonomy},
  author={Wu, Junyang and Luo, Mingyi and Xie, Fangfang and Zhang, Minghui and Zhang, Hanxiao and Zhang, Chunxi and Wang, Junhao and Sun, Jiayuan and Gu, Yun and Yang, Guang-Zhong},
  journal={arXiv preprint arXiv:2603.07909},
  year={2026}
}

@article{zhang2024robotic,
  title={Robotic-assisted bronchoscopy for the diagnosis of peripheral pulmonary lesions: A systematic review and meta-analysis},
  author={Zhang, Chunxi and Xie, Fangfang and Li, Runchang and Cui, Ningxin and Herth, Felix JF and Sun, Jiayuan},
  journal={Thoracic cancer},
  volume={15},
  number={7},
  pages={505--512},
  year={2024},
  publisher={Wiley Online Library}
}

@article{kops2023diagnostic,
  title={Diagnostic yield and safety of navigation bronchoscopy: a systematic review and meta-analysis},
  author={Kops, Stephan EP and Heus, Pauline and Korevaar, Dani{\"e}l A and Damen, Johanna AA and Idema, Demy L and Verhoeven, Roel LJ and Annema, Jouke T and Hooft, Lotty and van der Heijden, Erik HFM},
  journal={Lung cancer},
  volume={180},
  pages={107196},
  year={2023},
  publisher={Elsevier}
}

@article{abdelghani2024conebeam,
  title={Cone-beam computed tomography-guided shape-sensing robotic bronchoscopy vs. electromagnetic navigation bronchoscopy for pulmonary nodules},
  author={Abdelghani, Ramsy and Espinoza, Diana and Uribe, Juan P and Becnel, David and Herr, Rachel and Villalobos, Regina and Kheir, Fayez},
  journal={Journal of thoracic disease},
  volume={16},
  number={9},
  pages={5529--5538},
  year={2024}
}

@article{brohan2022rt,
  title={Rt-1: Robotics transformer for real-world control at scale},
  author={Brohan, Anthony and Brown, Noah and Carbajal, Justice and Chebotar, Yevgen and Dabis, Joseph and Finn, Chelsea and Gopalakrishnan, Keerthana and Hausman, Karol and Herzog, Alex and Hsu, Jasmine and others},
  journal={arXiv preprint arXiv:2212.06817},
  year={2022}
}

@inproceedings{Zhao-RSS-23,
  author    = {Tony Z. Zhao and Vikash Kumar and Sergey Levine and Chelsea Finn},
  title     = {Learning Fine-Grained Bimanual Manipulation with Low-Cost Hardware},
  booktitle = {Proceedings of Robotics: Science and Systems XIX},
  year      = {2023},
  address   = {Daegu, Republic of Korea},
  month     = {July},
  doi       = {10.15607/RSS.2023.XIX.016}
}

@article{chi2025diffusion,
  title={Diffusion policy: Visuomotor policy learning via action diffusion},
  author={Chi, Cheng and Xu, Zhenjia and Feng, Siyuan and Cousineau, Eric and Du, Yilun and Burchfiel, Benjamin and Tedrake, Russ and Song, Shuran},
  journal={The International Journal of Robotics Research},
  volume={44},
  number={10-11},
  pages={1684--1704},
  year={2025},
  publisher={Sage Publications, Sage UK: London, England}
}

@inproceedings{
kit2025endovla,
title={Endo{VLA}: Dual-Phase Vision-Language-Action for Precise Autonomous Tracking in Endoscopy},
author={CHI KIT NG and Long Bai and Guankun Wang and Yupeng Wang and Huxin Gao and Kun yuan and Chenhan Jin and Tieyong Zeng and Hongliang Ren},
booktitle={9th Annual Conference on Robot Learning},
year={2025},
url={https://openreview.net/forum?id=7XyO9Y1hI1}
}

@INPROCEEDINGS{11777392,
  author={Lin, Jinsong and Ng, Chi Kit and Xiong, Zhiyong and Pan, Zikang and Hu, Yihan and Tabassum, Tamima and Hao, Ziyi and Cheung, Eddie and Lai, Jiewen and Gao, Huxin and Ren, Hongliang},
  booktitle={2026 IEEE/RSJ International Conference on Intelligent Robots and Systems (IROS)}, 
  title={BiliVLA: Scene-Aware Vision-Language-Action Model with Reinforcement Learning for Autonomous Biliary Endoscopic Navigation}, 
  year={2026},
  volume={},
  number={},
  pages={13430-13437},
  doi={10.1109/IROS61738.2026.11777392}}

@inproceedings{sganga2019offsetnet,
  title={{OffsetNet}: Deep learning for localization in the lung using rendered images},
  author={Sganga, Jake and Eng, David and Graetzel, Chauncey and Camarillo, David},
  booktitle={2019 international conference on robotics and automation (ICRA)},
  pages={5046--5052},
  year={2019},
  organization={IEEE}
}

@article{zhang2024ai,
  title={AI co-pilot bronchoscope robot},
  author={Zhang, Jingyu and Liu, Lilu and Xiang, Pingyu and Fang, Qin and Nie, Xiuping and Ma, Honghai and Hu, Jian and Xiong, Rong and Wang, Yue and Lu, Haojian},
  journal={Nature communications},
  volume={15},
  number={1},
  pages={241},
  year={2024},
  publisher={Nature Publishing Group UK, London}
}

@article{bajcsy1988active,
  title={Active perception},
  author={Bajcsy, Ruzena},
  journal={Proceedings of the IEEE},
  volume={76},
  number={8},
  pages={966--1005},
  year={1988},
  publisher={IEEE}
}

@article{christiano2017deep,
  title={Deep reinforcement learning from human preferences},
  author={Christiano, Paul F and Leike, Jan and Brown, Tom and Martic, Miljan and Legg, Shane and Amodei, Dario},
  journal={Advances in neural information processing systems},
  volume={30},
  year={2017}
}

@inproceedings{brown2019extrapolating,
  title={Extrapolating beyond suboptimal demonstrations via inverse reinforcement learning from observations},
  author={Brown, Daniel and Goo, Wonjoon and Nagarajan, Prabhat and Niekum, Scott},
  booktitle={International conference on machine learning},
  pages={783--792},
  year={2019},
  organization={PMLR}
}

@inproceedings{ross2011reduction,
  title={A reduction of imitation learning and structured prediction to no-regret online learning},
  author={Ross, St{\'e}phane and Gordon, Geoffrey and Bagnell, Drew},
  booktitle={Proceedings of the fourteenth international conference on artificial intelligence and statistics},
  pages={627--635},
  year={2011},
  organization={JMLR Workshop and Conference Proceedings}
}

@inproceedings{tian2025jacobian,
  title={Jacobian Exploratory Dual-Phase Reinforcement Learning for Dynamic Endoluminal Navigation of Deformable Continuum Robots},
  author={Tian, Yu and Ng, Chi Kit and Ren, Hongliang},
  booktitle={2025 IEEE/RSJ International Conference on Intelligent Robots and Systems (IROS)},
  pages={21559--21565},
  year={2025},
  organization={IEEE}
}

@article{lin2026endowam,
  author = {Lin, Jinsong and Pan, Zikang and Liu, Wanhao and Ng, Chi Kit and Shao, Liangjing and Yu, Zihang and Wang, Ziyu and Wang, Yin and Wang, Jiaxi and Teoh, Jeremy Yuen-Chun and others},
  journal = {arXiv preprint arXiv:2608.01221},
  title = {EndoWAM: A Grounded World-Action Model for Generalizable Endoscopic Navigation},
  year = {2026}
}

@article{pomerleau1988alvinn,
  title={{ALVINN}: An autonomous land vehicle in a neural network},
  author={Pomerleau, Dean A},
  journal={Advances in neural information processing systems},
  volume={1},
  year={1988}
}

@article{ng2026endolift,
  title={EndoLIFT: Language-Disambiguated Latent-Conditioned Rectified Flow for Bidirectional Endoscopic Control},
  author={Ng, Chi Kit and Zhang, Yidong and Hing, Lui Siu and Lin, Jinsong and Wu, Tianchun and Chim, Ho Yin and Tang, Zhiqing and Yang, Tao and Gao, Huxin and Yeung, Trevor and others},
  journal={arXiv preprint arXiv:2608.20478},
  year={2026}
}

@inproceedings{ng2025contact,
  title={Contact-Aided Navigation of Flexible Robotic Endoscope Using Deep Reinforcement Learning in Dynamic Stomach},
  author={Ng, Chi Kit and Gao, Huxin and Ren, Tian-Ao and Lai, Jiewen and Ren, Hongliang},
  booktitle={2025 IEEE International Conference on Robotics and Biomimetics (ROBIO)},
  pages={797--803},
  year={2025},
  organization={IEEE}
}

@article{ng2026endovla,
  title={EndoVLA-OOV: Language-Conditioned Out-of-View Target Acquisition in Robotic Endoscopy},
  author={Ng, Chi Kit and Zhang, Yidong and Wu, Tianchun and Tang, Zhiqing and Zhao, Hanlin and Lui, Siu Hing and Yang, Tao and Lin, Jingsong and Gao, Huxin and Trevor, Yeung and others},
  journal={arXiv preprint arXiv:2610.24747},
  year={2026}
}

@article{ng2026endorsi,
  title={EndoRSI: Physical-Evidence-Driven Recursive Improvement of Endoscopic Robot Policies},
  author={Ng, Chi Kit and Zhang, Yidong and Wu, Tianchun and Lui, Siu Hing and Zhao, Hanlin and Gao, Huxin and Ren, Hongliang},
  journal={arXiv preprint arXiv:2610.13376},
  year={2026}
}

@article{giri2021virtual,
  title={Virtual bronchoscopic navigation versus non-virtual bronchoscopic navigation assisted bronchoscopy for the diagnosis of peripheral pulmonary lesions: a systematic review and meta-analysis},
  author={Giri, Mohan and Puri, Anju and Wang, Ting and Huang, Guichuan and Guo, Shuliang},
  journal={Therapeutic Advances in Respiratory Disease},
  volume={15},
  pages={17534666211017048},
  year={2021},
  publisher={SAGE Publications, Sage UK: London, England}
}

@article{kho2024bronchial,
  title={Bronchial branch tracing navigation in ultrathin bronchoscopy-guided radial endobronchial ultrasound for peripheral pulmonary nodule},
  author={Kho, Sze Shyang and Tan, Shirin Hui and Chan, Swee Kim and Chai, Chan Sin and Tie, Siew Teck},
  journal={BMC Pulmonary Medicine},
  volume={24},
  number={1},
  pages={466},
  year={2024},
  publisher={Springer}
}

@inproceedings{guo2025endo3r,
  title={{Endo3R}: Unified online reconstruction from dynamic monocular endoscopic video},
  author={Guo, Jiaxin and Dong, Wenzhen and Huang, Tianyu and Ding, Hao and Wang, Ziyi and Kuang, Haomin and Dou, Qi and Liu, Yun-Hui},
  booktitle={International Conference on Medical Image Computing and Computer-Assisted Intervention},
  pages={170--180},
  year={2025},
  organization={Springer}
}

@article{yang20243d,
  title={{3D} reconstruction from endoscopy images: A survey},
  author={Yang, Zhuoyue and Dai, Ju and Pan, Junjun},
  journal={Computers in Biology and Medicine},
  volume={175},
  pages={108546},
  year={2024},
  publisher={Elsevier}
}

@article{cold2025artificial,
  title={Artificial intelligence in bronchoscopy: a systematic review},
  author={Cold, Kristoffer Mazanti and Vamadevan, Anishan and Laursen, Christian B and Bjerrum, Flemming and Singh, Suveer and Konge, Lars},
  journal={European Respiratory Review},
  volume={34},
  number={176},
  pages={240274},
  year={2025},
  publisher={European Respiratory Society}
}
\end{document}